\documentclass{article}
\usepackage[preprint]{neurips_2026}

\usepackage{amsmath,amssymb,amsfonts}
\usepackage{booktabs}
\usepackage{multirow}
\usepackage{graphicx}
\usepackage{xcolor}
\usepackage{enumitem}
\usepackage{subcaption}
\usepackage{algorithm}
\usepackage{algpseudocode}
\usepackage{microtype}
\usepackage{etoolbox}
\usepackage{flafter}   
\PassOptionsToPackage{hyphens}{url}

\long\def\comment#1{}

\usepackage{hyperref}
\hypersetup{
  pdftitle={Closing the BF16--FP16 Convergence Speed Gap in RL~Post-Training with Stochastic Rounding},
  pdfauthor={Mahdi Nazemi},
  pdfsubject={Numerical formats in reinforcement learning post-training},
  pdfkeywords={stochastic rounding, BF16, FP16, reinforcement learning, RLVR, GRPO, weight sticking, mixed-precision training},
  colorlinks=true,
  linkcolor=red!72!yellow!71!black,
  citecolor=green!50!black,
  urlcolor=blue,
  bookmarksnumbered=true,
  pdfdisplaydoctitle=true,
  pdflang={en-US},
}
\usepackage[capitalize,nameinlink]{cleveref}

\bibpunct{[}{]}{,}{a}{,}{,}

\makeatletter
\def\bstctlcite{\@ifnextchar[{\@bstctlcite}{\@bstctlcite[@auxout]}}
\def\@bstctlcite[#1]#2{
  \@bsphack
  \@for\@citeb:=#2\do{
    \edef\@citeb{\expandafter\@firstofone\@citeb}
    \if@filesw
      \immediate\write\csname #1\endcsname{\string\citation{\@citeb}}
    \fi
  }
  \@esphack
}
\makeatother

\makeatletter
\patchcmd{\@maketitle}{Address \\}{}{}{
  \PackageWarning{paper}{Failed to remove anonymous address line from title block}
}
\makeatother

\makeatletter
  \renewcommand{\@notice}{}
\makeatother

\title{Closing the BF16--FP16 Convergence Speed Gap in RL~Post-Training with Stochastic Rounding}

\author{%
  Mahdi Nazemi \quad Oleg Rybakov \quad Tijmen Blankevoort \quad Asit Mishra \quad Carlo
  del Mundo\\
  NVIDIA \\
}

\begin{document}

\bstctlcite{BSTcontrol}

\maketitle

\begin{abstract}

BF16 is widely used to pretrain large language models because it has the same dynamic
range as FP32 and converges similarly to FP32 training.
However, in reinforcement learning with verifiable rewards, where learning rates are
typically low, we find that standard mixed-precision BF16 training converges more slowly
than FP16 training at the same learning rate.
We attribute this gap to \emph{weight sticking}: when FP32 shadow weights are quantized
to BF16 using round-to-nearest, small updates often leave the BF16 weights used for
rollout generation and training unchanged.
Consequently, several optimizer steps may be needed before updates to the shadow weights
are reflected in the BF16 weights.
Raising the BF16 learning rate reduces weight sticking and matches FP16's convergence
speed, but it also increases the KL divergence from the reference model.
We instead apply stochastic rounding when quantizing the FP32 shadow weights to BF16.
This approach matches FP16's convergence speed across different models and datasets
while reducing the KL divergence relative to BF16 training with a larger learning rate.
Unlike FP16, it keeps BF16's wider dynamic range and needs no loss scaling.

\end{abstract}

\begin{figure}[!b]
\centering
\includegraphics[width=\linewidth]{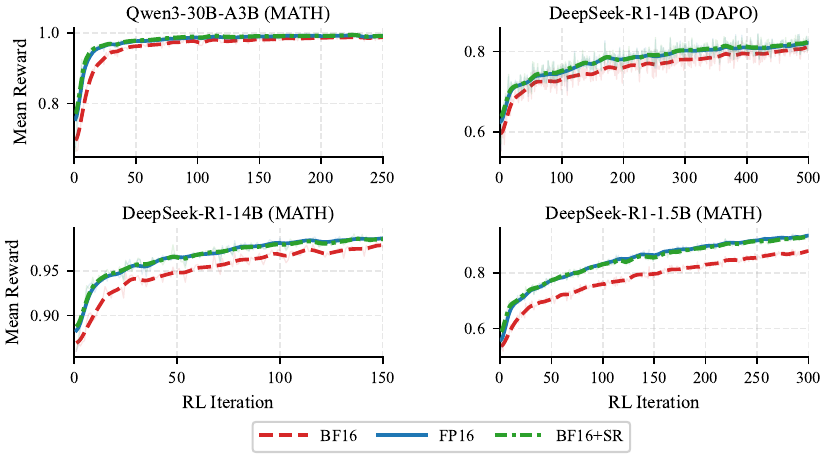}
\caption{Applying stochastic rounding when quantizing the FP32 shadow weights to BF16
(BF16+SR) closes the BF16--FP16 convergence speed gap across different models and
datasets.
The FP16 and BF16+SR reward trajectories closely overlap.
}
\label{fig:teaser}
\end{figure}

\section{Introduction}
\label{sec:intro}

Mixed-precision BF16 training is widely used for pretraining large language
models~\citep{DBLP:journals/corr/abs-2407-21783,DBLP:journals/corr/abs-2401-02954}.
BF16 has the same dynamic range as FP32, and mixed-precision BF16 training converges
similarly to FP32 training~\citep{DBLP:journals/corr/abs-1905-12322}.
BF16 is also broadly supported by modern AI accelerators, which typically offer
substantially higher peak throughput for BF16 matrix multiplications than for FP32.
These properties make BF16 a natural choice for reinforcement learning (RL)
post-training as well.

However, at the low learning rates (LRs) common in reinforcement learning with
verifiable rewards (RLVR), such as \( 10^{-6}
\)~\citep{DBLP:conf/nips/YuZZYZYDFLLLLLL25,DBLP:journals/corr/abs-2503-20783}, we find
that BF16 needs more RL iterations than FP16 to reach the same reward or downstream task
score.
This gap persists even when the rollout and training backends produce bit-exact
log-probabilities for the same weights and inputs, so training-inference
mismatch~\citep{DBLP:journals/corr/abs-2510-26788} alone cannot explain it.
Throughout this paper, we measure convergence speed by the number of RL iterations
needed to reach a given reward or score.
We attribute the slower convergence to weight sticking: when FP32 shadow weights are
quantized to BF16 using round-to-nearest, small updates often leave the BF16 weights
used for rollout generation and training unchanged for several optimizer steps.
Formats with fewer mantissa bits, such as BF16, are more prone to
sticking~\citep{DBLP:journals/corr/abs-2602-03839,dirhoussi2026_defeating_the_trainer_generator_precision_mismatch_in_trl}.

Two simple fixes can reduce weight sticking: switching to FP16 or increasing the BF16
LR.
Because of its narrower dynamic range, FP16 typically requires loss scaling to avoid
gradient underflow.
In preliminary FP16 RLVR experiments on agentic tasks, we observed frequent overflows
even with loss scales much smaller than those common in pretraining.
Each overflow skips the corresponding optimizer step, so the rollouts used for that step
are wasted.
Increasing the BF16 LR can recover FP16's convergence speed, but it also increases the
policy's drift from the reference model, which we measure by KL divergence.
Such drift is commonly penalized in RL
post-training~\citep{DBLP:journals/corr/abs-2512-02556}, and it has been shown to
predict how much a post-trained model forgets~\citep{DBLP:journals/corr/abs-2509-04259}.

We instead apply stochastic rounding (SR) when quantizing the FP32 shadow weights to
BF16.
SR rounds each shadow weight to one of its neighboring BF16 values, with probabilities
set by the shadow weight's position between them.
Even a small update to a shadow weight therefore changes the rounding probabilities, so
the expected value of each BF16 weight moves with its FP32 shadow weight.
BF16 with SR (BF16+SR) matches FP16's convergence speed (\cref{fig:teaser}) with less
drift than BF16 training at a higher LR.
It also benefits from BF16's broad accelerator support and avoids the need for FP16 loss
scaling.

Our contributions are as follows:
\begin{itemize}

\item We show that, at typical RLVR LRs, BF16 converges more slowly than FP16 even when
      the rollout and training backends produce bit-exact log-probabilities for the same
      weights and inputs.

\item We identify weight sticking as a contributor to this slower convergence and show
      that increasing the number of mantissa bits in simulated BF16-like formats speeds
      up convergence.

\item We show that applying SR when quantizing the FP32 shadow weights to BF16 matches
      FP16's convergence speed with less drift than raising the BF16 LR, and that
      applying SR in the forward passes alone is sufficient.

\end{itemize}

\section{Weight Sticking and Stochastic Rounding}
\label{sec:weight-sticking-sr}

Standard mixed-precision training applies optimizer updates to FP32 shadow weights and
quantizes weights to lower-precision formats, such as BF16 or FP16, for the forward and
backward passes~\citep{DBLP:conf/iclr/MicikeviciusNAD18}.
In RL post-training, the rollout backend also uses the low-precision weights, so this
weight quantization affects the policy that produces training data.

Let \( w \in \mathbb{R} \) denote an FP32 shadow weight, and let \( w^- \) and \( w^+ \)
be adjacent representable values in the target format such that \( w^- \leq w < w^+ \).
Their spacing, \( \mathrm{ULP}(w) = w^+ - w^- \), is one unit in the last place (ULP) of
the target format at \( w \).
We consider quantizers \( Q \) that map \( w \) to either \( w^- \) or \( w^+ \).
Round-to-nearest (RN) maps \( w \) to
\begin{equation}
\label{eq:rn}
Q^{\mathrm{RN}}(w) =
\begin{cases}
w^-,            & w < \frac{w^- + w^+}{2}, \\
w^+,            & w > \frac{w^- + w^+}{2}, \\
w_{\text{tie}}, & w = \frac{w^- + w^+}{2},
\end{cases}
\end{equation}
where \( w_{\text{tie}} \in \{w^-, w^+\} \) depends on the tie-breaking rule.

Under RN quantization, weight sticking occurs when a nonzero update \( u \) to an FP32
shadow weight \( w \) leaves its quantized value unchanged, i.e., \( Q^{\mathrm{RN}}(w +
u) = Q^{\mathrm{RN}}(w) \).
Here, \( u \) may be a single optimizer update or the net update over several
consecutive steps.

When post-training starts from a BF16 checkpoint, each initial shadow weight \( w_0 \)
lies on the BF16 grid, and, ignoring special cases, its quantized value stays at \( w_0
\) as long as the net update is smaller than half a ULP:
\begin{equation}
\label{eq:sticking}
|u| < \frac{\mathrm{ULP}(w_0)}{2} \implies Q^{\mathrm{RN}}(w_0 + u) = Q^{\mathrm{RN}}(w_0).
\end{equation}
BF16 has 7 mantissa bits compared to FP16's 10, so for values in the normal range of
both formats, the BF16 ULP is \( 8 \times \) the FP16 ULP, and a BF16 weight needs an \(
8 \times \) larger net update to move.
Equivalently, since the ULP grows with \( |w_0| \), a given update can move FP16 weights
up to about \( 8 \times \) larger in magnitude than the largest BF16 weights it can
move.

For a stuck weight, the RN quantization error exactly cancels the optimizer's net update
to \( w_0 \):
\begin{equation}
\label{eq:rn-error}
e^{\mathrm{RN}}(w_0 + u) = Q^{\mathrm{RN}}(w_0 + u) - (w_0 + u) = -u.
\end{equation}
To remove this systematic error in expectation, SR quantizes a shadow weight according
to
\begin{equation}
\label{eq:sr}
Q^{\mathrm{SR}}(w)=
\begin{cases}
w^+ & \text{with probability } \displaystyle \frac{w - w^-}{w^+ - w^-}, \\
w^- & \text{with probability } \displaystyle \frac{w^+ - w}{w^+ - w^-}.
\end{cases}
\end{equation}
Consequently, \( \mathbb{E} \! \left[ Q^{\mathrm{SR}}(w_0 + u) \mid w_0 + u \right] =
w_0 + u \), where the expectation is over the rounding randomness.
SR replaces RN's systematic cancellation of updates in stuck weights with a rounding
error that has zero mean but higher variance.
We hypothesize that the added variance harms RLVR convergence less than RN's systematic
error does, even when rollout generation and training use a single random realization of
the quantized weights rather than their expectation.
We test this hypothesis in \cref{sec:experiments} by applying SR when quantizing the
FP32 shadow weights to BF16 while leaving the rest of the training recipe unchanged.

\section{Experiments}
\label{sec:experiments}

\subsection{Experimental Setup}
\label{sec:exp:setup}

\paragraph{Models and datasets.}

Most experiments train DeepSeek-R1-Distill-Qwen-1.5B
(DeepSeek-R1-1.5B)~\citep{DBLP:journals/corr/abs-2501-12948} on
MATH-1460~\citep{DBLP:journals/corr/abs-2510-26788}, a curated set of 1,460 competition
math problems with binary correctness rewards.
Larger-scale experiments train DeepSeek-R1-Distill-Qwen-14B
(DeepSeek-R1-14B)~\citep{DBLP:journals/corr/abs-2501-12948} on MATH-1460 or
DAPO-Math-17k~\citep{DBLP:conf/nips/YuZZYZYDFLLLLLL25}, and
Qwen3-30B-A3B~\citep{DBLP:journals/corr/abs-2505-09388} on MATH-1460.
The TorchInductor study in \cref{sec:exp:inductor} also trains
Qwen3-1.7B~\citep{DBLP:journals/corr/abs-2505-09388} on MATH-1460.
We measure downstream accuracy on the AIME 2024 and AIME 2025 datasets and report
mean@32, the accuracy averaged over 32 sampled responses per problem.

\paragraph{Training configuration.}

Unless noted otherwise, experiments use
Dr.~GRPO~\citep{DBLP:journals/corr/abs-2503-20783} in the verl
framework~\citep{DBLP:conf/eurosys/ShengZYWZZPL025}, with
vLLM~0.12~\citep{DBLP:conf/sosp/KwonLZ0ZY0ZS23} for rollout generation and Hugging~Face
Transformers~\citep{DBLP:conf/emnlp/WolfDSCDMCRLFDS20} as the training backend.
We use asymmetric clip ratios \( \epsilon_\text{lo}=0.2 \) and \(
\epsilon_\text{hi}=0.28 \)~\citep{DBLP:conf/nips/YuZZYZYDFLLLLLL25}, a dual-clip lower
bound \( c=3.0 \)~\citep{DBLP:conf/aaai/YeLSSZWYYWGCYZS20}, and a constant token-budget
loss normalizer~\citep{DBLP:journals/corr/abs-2503-20783}.
We use AdamW with FP32 shadow weights and a default LR of \( \eta = 10^{-6} \).
Each RL iteration samples 8~responses for each of 512~problems and splits the resulting
4,096~sequences into 4~mini-batches of 1,024, so each RL iteration takes 4~optimizer
steps.
The maximum response length is 8,192~tokens.
FP16 runs use dynamic loss scaling with PyTorch's \texttt{ShardedGradScaler} and a
growth interval of 400~steps.

\paragraph{Numerical parity and hardware.}

For tightly controlled experiments, we add run-to-run determinism and training-inference
numerical parity for selected models in verl.
Determinism makes runs reproducible, and parity removes training-inference mismatch as a
confounder, so numerical formats and rounding schemes can be compared fairly.
We achieve parity by using batch-invariant kernels~\citep{he2025nondeterminism} and by
aligning the numerics of the rollout and training backends so that they produce
bit-exact log-probabilities for the same weights and inputs.
Enforcing parity reduces throughput, so we reserve it for the BF16--FP16 parity
comparison and the weight sticking measurements in \cref{sec:exp:results}, which run on
64 NVIDIA H100 GPUs.
All other experiments run on 64 NVIDIA B200 GPUs.

\paragraph{Data types.}

Regardless of the numerical format or rounding scheme, attention accumulation, RMS
normalization, log-softmax, optimizer states, and gradient all-reduce use FP32.
Matrix multiplications and their inputs (weights, activations, and gradients), attention
outputs, and residual additions use BF16 or FP16.

\subsection{Results and Discussion}
\label{sec:exp:results}

\paragraph{BF16 converges more slowly than FP16 even after removing training-inference mismatch.}
\label{sec:exp:inductor}
\label{sec:exp:parity}

\begin{figure}[tb]
\centering
\includegraphics[width=\linewidth]{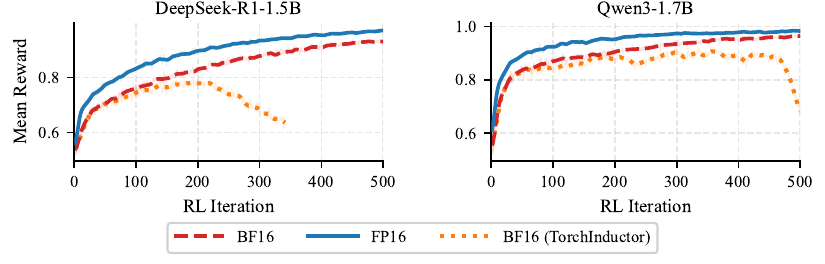}
\caption{Disabling TorchInductor compiler optimizations in vLLM prevents BF16 reward
collapse.}
\label{fig:inductor_collapse}
\end{figure}

\begin{figure}[tb]
\centering
\includegraphics[width=\linewidth]{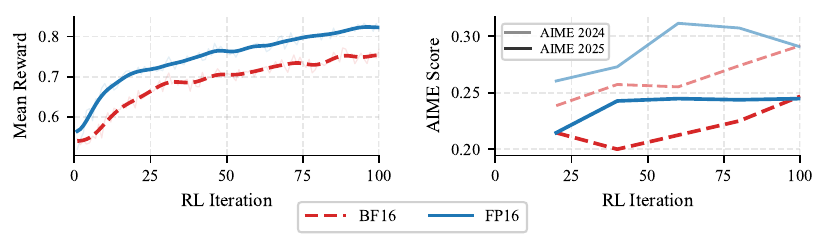}
\caption{Under training-inference numerical parity, FP16 reaches a given mean reward and
AIME score (mean@32) in fewer RL iterations than BF16 (DeepSeek-R1-1.5B, MATH-1460).
}
\label{fig:parity_gap}
\end{figure}

\citet{DBLP:journals/corr/abs-2510-26788} reported reward collapse in BF16 RLVR and
attributed it to training-inference mismatch.
In our setup, the collapse occurs only when vLLM's TorchInductor compiler optimizations
are enabled; disabling them prevents it (\cref{fig:inductor_collapse}) and increases
end-to-end step time by less than 2\%.
We observed this behavior with DeepSeek-R1-1.5B and Qwen3-1.7B on all vLLM releases we
used during this work (0.12 to 0.17), but we have not tested later releases (the latest
is 0.33 at the time of writing).
We therefore disable TorchInductor optimizations in all remaining experiments.
With TorchInductor disabled and parity enforced, FP16 still reaches a given reward level
in fewer RL iterations than BF16 at the same LR (\cref{fig:parity_gap}).
In the FP16 run, \texttt{ShardedGradScaler} never skipped an optimizer step.

\paragraph{BF16 has a higher weight sticking rate than FP16 at RLVR's low LRs.}
\label{sec:exp:weight_sticking}

\Cref{tab:sticking} reports the percentage of weights that still equal their values in
the BF16 checkpoint after each of the four optimizer steps in the first RL iteration.
Across the transformer blocks of the DeepSeek-R1-1.5B model, the median weight magnitude
is \( 2.5 \times 10^{-2} \), and the 25th and 75th percentiles are \( 1.2 \times 10^{-2}
\) and \( 4.5 \times 10^{-2} \).
At the median magnitude, half a ULP is about \( 6.1 \times 10^{-5} \) in BF16 and \( 7.6
\times 10^{-6} \) in FP16, both of which are well above the Adam update of roughly \(
\eta = 10^{-6} \).
Most weights therefore remain unchanged in both formats.
However, because FP16's half-ULP threshold is \( 8 \times \) smaller, the same update
changes a larger fraction of the small-magnitude weights, and FP16 changes roughly \( 8
\times \) as many weights as BF16 by the end of the first RL iteration (9.21\% versus
1.16\%).

\begin{table}[tb]
\centering
\caption{Percentage of quantized weights equal to their initial checkpoint values after
each optimizer step in the first RL iteration (DeepSeek-R1-1.5B, MATH-1460, \( \eta =
10^{-6} \)).}
\label{tab:sticking}
\small
\begin{tabular}{@{}ccc@{}}
\toprule
Optimizer Step & BF16 & FP16 \\
\midrule
1              & 99.42\%    & 95.35\%    \\
2              & 99.23\%    & 93.86\%    \\
3              & 98.93\%    & 91.51\%    \\
4              & 98.84\%    & 90.79\%    \\
\bottomrule
\end{tabular}
\end{table}

\begin{figure}[tb]
\centering
\begin{subfigure}[b]{0.49\linewidth}
\centering
\includegraphics[width=\linewidth]{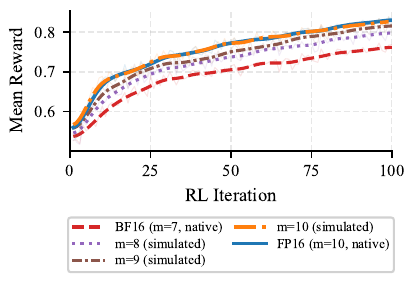}
\caption{Mantissa bit-width sweep}
\label{fig:sticking_mantissa}
\end{subfigure}
\hfill
\begin{subfigure}[b]{0.49\linewidth}
\centering
\includegraphics[width=\linewidth]{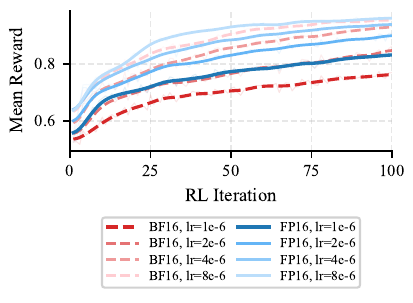}
\caption{LR sweep}
\label{fig:sticking_lr}
\end{subfigure}
\caption{Effect of mantissa bit-width and LR on convergence (DeepSeek-R1-1.5B,
MATH-1460).
(a) A simulated BF16-like format with 10 mantissa bits tracks FP16. (b) A higher LR speeds
up convergence for both formats, but FP16 stays ahead at every LR.
}
\label{fig:sticking_tests}
\end{figure}

\Cref{eq:sticking} suggests two ways to reduce weight sticking: shrinking the ULP, e.g.,
by increasing the number of mantissa bits, and increasing the optimizer's update size,
e.g., by increasing the LR.
\Cref{fig:sticking_mantissa} tests the first with simulated formats that keep BF16's 8
exponent bits but have \( m \in \{8, 9, 10\} \) mantissa bits.
Each added mantissa bit speeds up convergence, and the \( m = 10 \) format tracks FP16
closely throughout training.
These results support limited mantissa precision as a contributor to the convergence
speed gap.
\Cref{fig:sticking_lr} tests the second with LRs from \( 10^{-6} \) to \( 8 \times
10^{-6} \).
A higher LR speeds up convergence for both formats, but FP16 stays ahead of BF16 at
every LR.

\begin{figure}[tb]
\centering
\includegraphics[width=\linewidth]{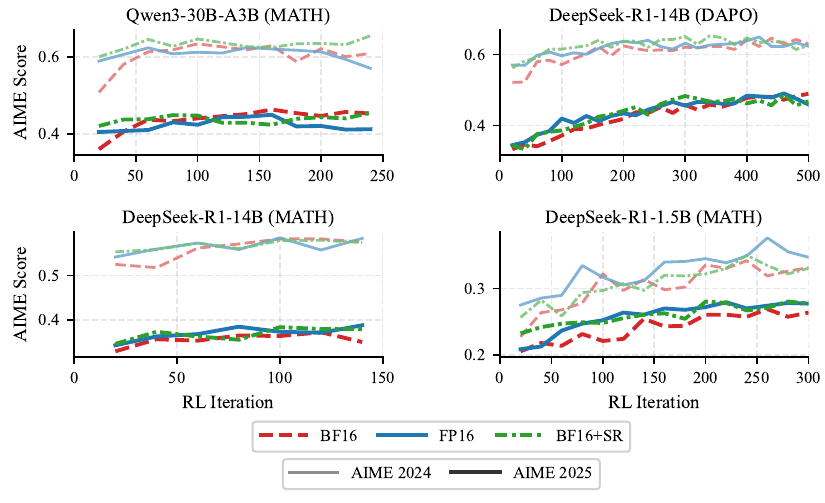}
\caption{AIME 2024 and AIME 2025 scores (mean@32).
BF16 initially trails FP16 and BF16+SR on downstream task scores but usually closes the
gap as training progresses.
}
\label{fig:evals}
\end{figure}

\begin{figure}[tb]
\centering
\includegraphics[width=\linewidth]{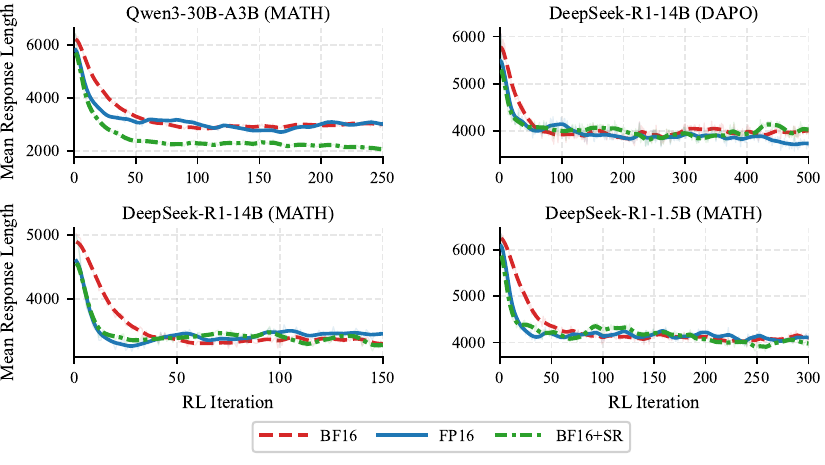}
\caption{Mean response length during training.
After an initial phase in which FP16 and BF16+SR produce shorter responses, all three
configurations converge to similar lengths, except on Qwen3-30B-A3B.
}
\label{fig:response_length}
\end{figure}

\begin{figure}[tb]
\centering
\includegraphics[width=\linewidth]{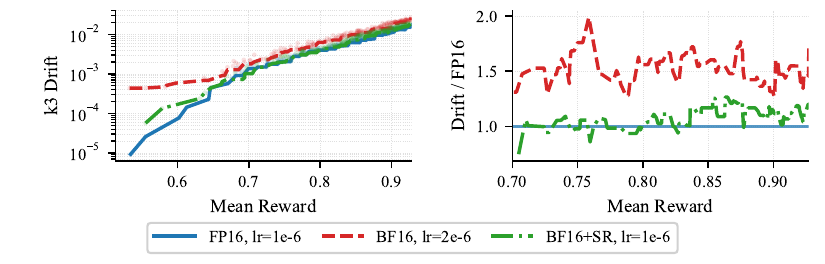}
\caption{Drift from the reference model at matched reward (DeepSeek-R1-1.5B, MATH-1460).
BF16 with a higher LR closes the reward gap but incurs larger drift, while BF16+SR at
the default LR stays close to FP16.
}
\label{fig:lr_vs_sr_drift}
\end{figure}

\paragraph{Stochastic rounding closes the convergence speed gap.}
\label{sec:exp:sr}

BF16+SR matches the FP16 reward trajectory across the evaluated model–dataset
combinations, spanning 1.5B, 14B, and 30B-A3B models and two training datasets
(\cref{fig:teaser}).
On AIME 2024 and AIME 2025, BF16 initially trails FP16 and BF16+SR but usually closes
the gap later in training (\cref{fig:evals}).
Early in training, FP16 and BF16+SR produce shorter responses than BF16 for all models
and datasets, but all three configurations eventually converge to similar response
lengths (\cref{fig:response_length}).
The only exception is the Qwen3-30B-A3B model, for which BF16+SR produces responses
about 25\% shorter than the other two configurations.
We leave the study of this behavior to future work.

\paragraph{A higher BF16 LR closes the gap but increases drift.}
\label{sec:exp:lr_drift}

BF16 at \( \eta = 2 \times 10^{-6} \) follows a reward trajectory similar to FP16 at \(
\eta = 10^{-6} \) (\cref{fig:sticking_lr}), so tuning the LR is a viable way to close
the gap.
However, a higher LR also changes the optimization dynamics, so we compare the resulting
drift from the reference model using the k3 estimator of the KL divergence.
\Cref{fig:lr_vs_sr_drift} plots the lowest measured drift at each reward level.
For mean rewards above 0.7, BF16 at \( \eta = 2 \times 10^{-6} \) drifts about \( 1.5
\times \) as much as FP16, whereas BF16+SR stays close to FP16.
Higher LRs can therefore close the convergence speed gap, but at the cost of more drift
from the reference model.

\paragraph{Stochastic rounding of forward-pass weights is sufficient.}
\label{sec:exp:sr_variants}

We compared several SR variants on DeepSeek-R1-1.5B and MATH-1460.
Drawing a fresh SR realization in every forward pass gives the same reward trajectory as
drawing one realization after each optimizer step and using it for all forward passes
until the next optimizer step, including rollout generation after weight
synchronization.
Also applying SR to the weights in the backward pass, or to the gradients, does not
speed up convergence further.
Applying SR only in the rollout backend, however, is not enough: the mean reward
plateaus at a value well below FP16's.
We therefore recommend the simplest variant: apply SR once after each optimizer step,
use the resulting BF16 weights for all forward passes in both backends until the next
optimizer step, and use round-to-nearest for all other conversions.

\section{Conclusions and Future Work}
\label{sec:conclusion}

We showed that, at the low LRs typical of RLVR, BF16 converges more slowly than FP16,
and that this convergence speed gap persists under training-inference numerical parity.
We attribute the gap to weight sticking: with RN and only 7 mantissa bits, a BF16 weight
reflects small updates to its shadow weight only after the net update crosses a rounding
boundary.

We compared three remedies.
Switching to FP16 works well for the models and datasets in this paper, but in
preliminary RLVR experiments with Qwen3.5-397B-A17B on an agentic software engineering
task, FP16 training overflowed frequently even at a loss scale of 1,024, which is much
smaller than the loss scales common in pretraining.
Raising the BF16 LR matches FP16's convergence speed but increases drift from the
reference model, which has been shown to predict how much a post-trained model forgets.
Applying stochastic rounding when quantizing the FP32 shadow weights to BF16 closes the
convergence speed gap with drift close to FP16's.

We expect weight sticking to worsen in formats with even fewer mantissa bits than BF16.
AI accelerators typically provide higher throughput for matrix multiplications in
lower-precision formats, so weight sticking could limit their use for RLVR.
Preliminary asynchronous RLVR experiments on Qwen3.5-397B-A17B, in which rollouts lag
the trained policy by at most four policy versions, suggest that SR speeds up
convergence in BF16 as well as in low-precision formats such as MXFP8.
We leave a detailed study of RLVR using low-precision formats to future work.

\clearpage
{\small
\bibliographystyle{IEEEtranSN}
\bibliography{references}
}

\end{document}